# Visual-Inertial Odometry for Ground Robots with Dynamic Camera-IMU Extrinsic Calibration

Bo Dong

*Abstract*—**Visual-inertial odometry systems are often constrained by the assumption that sensors are rigidly mounted, which limits their adaptability in practical scenarios involving structural motion. To address this limitation, this paper investigates time-varying camera-IMU extrinsic calibration. First, a dynamic camera-IMU extrinsic calibration method based on the Denavit-Hartenberg convention is proposed by incorporating the robot kinematic model. Second, a two-step initialization strategy is developed to efficiently estimate the initial joint angles, providing a reliable starting point for subsequent extrinsic parameter optimization and enabling online estimation of time-varying camera-IMU extrinsic parameters. Simulation experiments demonstrate that the proposed framework can effectively accommodate variations in camera-IMU extrinsic parameters while maintaining competitive localization accuracy.**

## I. Introduction

Visual-inertial odometry (VIO) integrates camera observations with inertial measurements from an inertial measurement unit (IMU) to overcome the scale ambiguity of monocular visual odometry and improve motion estimation robustness. By exploiting the complementary characteristics of visual and inertial sensors, VIO enables metric-scale motion estimation and provides improved localization accuracy under challenging environmental conditions.

The extrinsic calibration between the camera and the IMU, including both rotational and translational components, plays a critical role in the performance of VIO systems. Inaccurate extrinsic parameters can introduce inconsistent multi-sensor measurements, resulting in localization drift and degraded robustness. Consequently, numerous studies have investigated camera-IMU extrinsic calibration through either offline calibration or online self-calibration methods [1], [2], [3]. However, most existing approaches assume that the camera and the IMU are rigidly mounted, such that the extrinsic parameters remain constant during system operation. In many practical ground robot applications, the sensing platform is mounted on movable mechanisms or articulated structures to expand the field of view, support active perception, or satisfy platform-specific mechanical constraints. As a result, the relative transformation between the camera and the IMU varies continuously with robot motion, violating the rigid extrinsic assumption adopted by conventional VIO systems. Under these conditions, treating the extrinsic parameters as constants may introduce systematic errors and degrade localization performance. A straightforward solution is to estimate the time-varying extrinsic parameters directly during optimization. However, dynamic extrinsic calibration is inherently challenging because the extrinsic parameters are strongly coupled with the system states, including the platform pose and sensor biases. Maintaining consistency between the evolving multi-sensor states and the dynamic camera-IMU transformation substantially increases the complexity of state estimation and makes accurate and robust online calibration difficult.

Bo Dong is with Tsinghua Shenzhen International Graduate School, Tsinghua University, Shenzhen 518055, China (e-mail: db19@tsinghua.org.cn).

The main contributions of this paper are summarized as follows:

- We propose a D-H-based kinematic modeling method for online estimation of time-varying camera-IMU extrinsic parameters in ground robot visual-inertial odometry, enabling dynamic extrinsic calibration by incorporating robot kinematic constraints.
- We analyze the relationship between the two dynamic joint angles and the logarithmic rotation residual cost function based on the D-H model, and develop a two-step initialization strategy to efficiently estimate their initial values for subsequent extrinsic parameter optimization.

## II. Related Works

[4] proposes an extrinsic calibration method for dynamic multi-camera clusters, where one or more cameras are mounted on actuated mechanisms. The inter-camera extrinsics are parameterized using the D-H representation, enabling the online estimation of time-varying transformations. Building on this work, [5] introduces an active calibration scheme to select viewpoints that maximize information gain, achieving high accuracy with significantly reduced data requirements. Unlike methods focused solely on multi-camera systems, [6] and [7] extend dynamic extrinsic calibration to camera-IMU systems on aerial platforms. By integrating time-varying extrinsic estimation into established optimization-based VIO frameworks, OKVIS [8] and VINS-Fusion [9], these approaches enable simultaneous localization and online parameter refinement. Furthermore, [7] provides a detailed degeneracy analysis for dynamic extrinsic calibration. The results indicate that, under sufficient excitation, the system can achieve localization accuracy comparable to conventional VIO systems. Nevertheless, these approaches primarily focus on calibration accuracy and do not explicitly address the

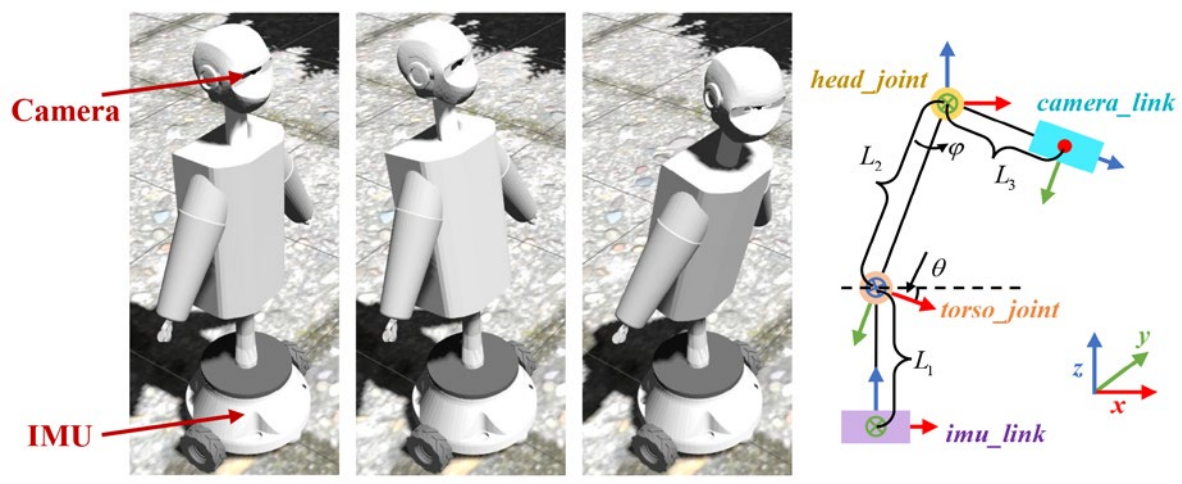

Fig. 1. Gazebo model of the wheeled humanoid robot. The robot supports torso pitch and head yaw motions. The first three images show three representative robot postures, while the fourth image shows a side view of the D-H coordinate frames along the kinematic chain from the IMU to the camera.

computational overhead introduced by online dynamic extrinsic estimation.

## III. Preliminary

The D-H convention is widely used in robotic systems to describe the relative transformations between consecutive links in articulated structures. Although frames can, in principle, be assigned arbitrarily to each link, the motion between adjacent links is typically constrained by specific joint mechanisms rather than being arbitrary. Therefore, defining link frames according to a consistent set of rules greatly facilitates system modeling and kinematic analysis. The D-H convention is specifically designed to provide such a systematic parameterization.

For two consecutive frames, denoted as $\mathcal{F}_{i-1}$ and $\mathcal{F}_i$, the transformation is described by four parameters: the link length $a_{i-1}$, link offset $d_i$, link twist $\alpha_{i-1}$, and joint angle $\theta_i$. Based on these parameters, the homogeneous transformation matrix from frame $\mathcal{F}_i$ to $\mathcal{F}_{i-1}$ can be expressed as:

$$\mathbf{T}_i^{i-1} = \begin{bmatrix} c_{\theta_i} & -s_{\theta_i} & 0 & a_{i-1} \\ s_{\theta_i} c_{\alpha_{i-1}} & c_{\theta_i} c_{\alpha_{i-1}} & -s_{\alpha_{i-1}} & -d_i s_{\alpha_{i-1}} \\ s_{\theta_i} s_{\alpha_{i-1}} & c_{\theta_i} s_{\alpha_{i-1}} & c_{\alpha_{i-1}} & d_i c_{\alpha_{i-1}} \\ 0 & 0 & 0 & 1 \end{bmatrix}, \tag{1}$$

where $c_{\cdot} = \cos(\bullet)$ and $s_{\cdot} = \sin(\bullet)$.

In articulated robotic systems, the kinematic chain from the base to the end-effector typically consists of multiple links. By assigning coordinate frames to consecutive links and parameterizing their relative transformations using the D-H convention, the transformation from the base frame to the end-effector can be obtained through successive composition of transformations. This representation provides a compact and structured formulation of the kinematic chain.

## IV. Methodology

### A. Kinematic Modeling of Dynamic Camera-IMU Extrinsics

In this work, a differential-drive wheeled humanoid robot is used as the experimental platform. Wheeled humanoid robots have potential applications in logistics, material transportation, and palletizing, where articulated mechanisms can introduce time-varying sensor configurations. As illustrated in Fig. 1, the robot is equipped with a monocular camera mounted in the head region and an IMU installed at the center of the mobile base. In this study, the articulated motions considered are limited to the torso and head joints, which provide a controlled platform for investigating the effect of time-varying camera-IMU extrinsic parameters on ground robot localization.

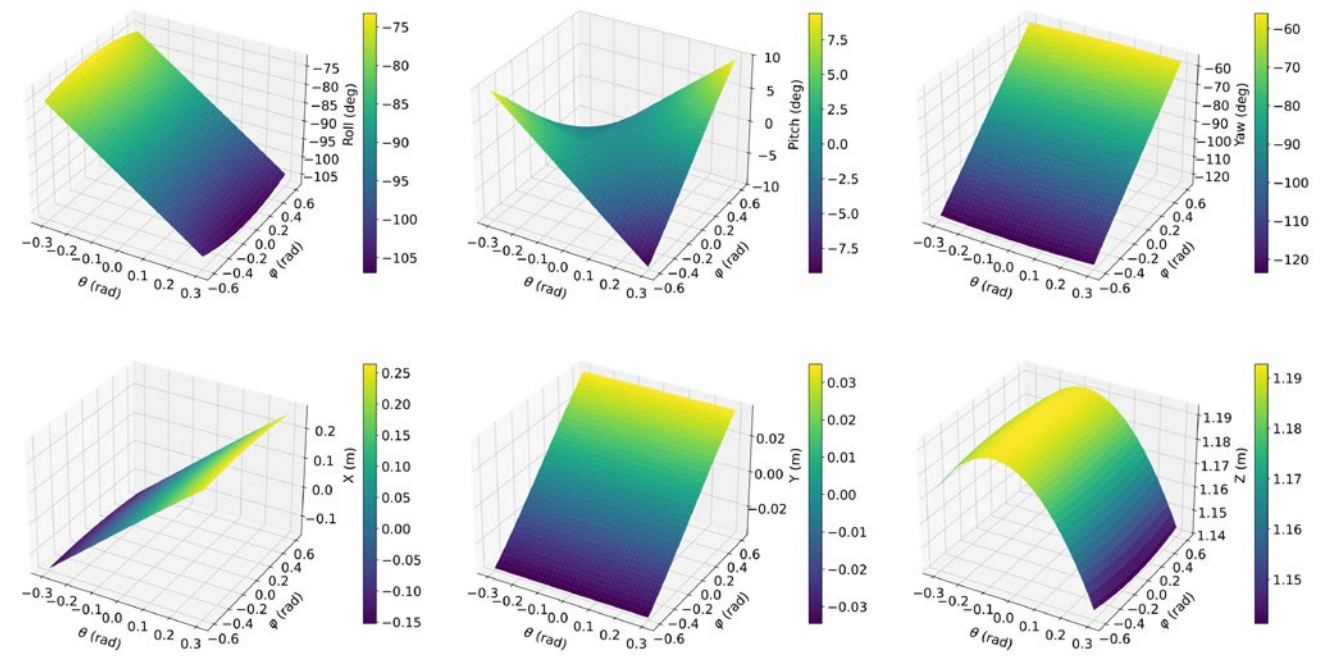

Fig. 2. Variation of the camera-IMU extrinsic parameters with respect to the torso pitch angle $\theta$ and head yaw angle $\varphi$ based on the proposed D-H kinematic model.

Unlike conventional wheeled robots with rigidly mounted sensors, the relative pose between the camera and the IMU in our platform is continuously time-varying during operation due to the articulated upper-body motion driven by the *torso_joint* and *head_joint*. To mathematically formulate this dynamic spatial relationship, the standard D-H convention is adopted to construct the kinematic chain between the IMU (*imu_link*), which is located at the center of the mobile base, and the camera optical frame (*camera_link*), as shown in Fig. 1. Specifically, this camera-IMU extrinsic transformation is determined by five parameters: two joint angles, $\theta$ and $\varphi$, representing the torso pitch and head yaw angles, respectively, and three constant mechanical parameters, $L_1$, $L_2$, and $L_3$, representing the base-to-torso, torso-to-head, and head-to-camera distances, respectively. Therefore, the camera-IMU extrinsic transformation can be expressed as

$$\begin{aligned} \mathbf{T}_C^I(\varphi,\theta) &= \begin{bmatrix} s_\varphi c_\theta & -s_\theta & c_\varphi c_\theta & L_2 s_\theta + L_3 c_\varphi c_\theta \\ -c_\varphi & 0 & s_\varphi & L_3 s_\varphi \\ -s_\varphi s_\theta & -c_\theta & -s_\theta c_\varphi & L_1 + L_2 c_\theta - L_3 s_\theta c_\varphi \\ 0 & 0 & 0 & 1 \end{bmatrix} \\ &= \begin{bmatrix} \mathbf{R}_C^I(\varphi,\theta) & \mathbf{t}_C^I(\varphi,\theta) \\ \mathbf{0}^T & 1 \end{bmatrix}, \end{aligned} \tag{2}$$

where $\mathbf{R}_C^I(\varphi,\theta) \in SO(3)$ and $\mathbf{t}_C^I(\varphi,\theta) \in \mathbb{R}^3$ represent the rotation and translation components of the extrinsic parameters, respectively. It can be observed from (2) that the time-varying component of the camera-IMU extrinsic parameters is solely determined by the two joint state $\theta$ and $\varphi$, whereas $L_1$, $L_2$, and $L_3$ are constant mechanical parameters.

To further analyze the relationship between the joint states and the camera-IMU extrinsic parameters, Fig. 2 illustrates the variation of the rotation and translation components with respect to $\theta$ and $\varphi$, which are directly derived from (2). It can be observed that the rotation components exhibit coupled variations with both joint angles, while different translation components show distinct dependencies. Specifically, the *y*-translation is mainly affected by the head yaw angle $\varphi$, whereas the *z*-translation presents a nonlinear variation due to the trigonometric relationships in the D-H formulation. These results demonstrate that the dynamic camera-IMU extrinsic parameters are constrained by the robot kinematic structure and can be represented as functions of only two joint states.

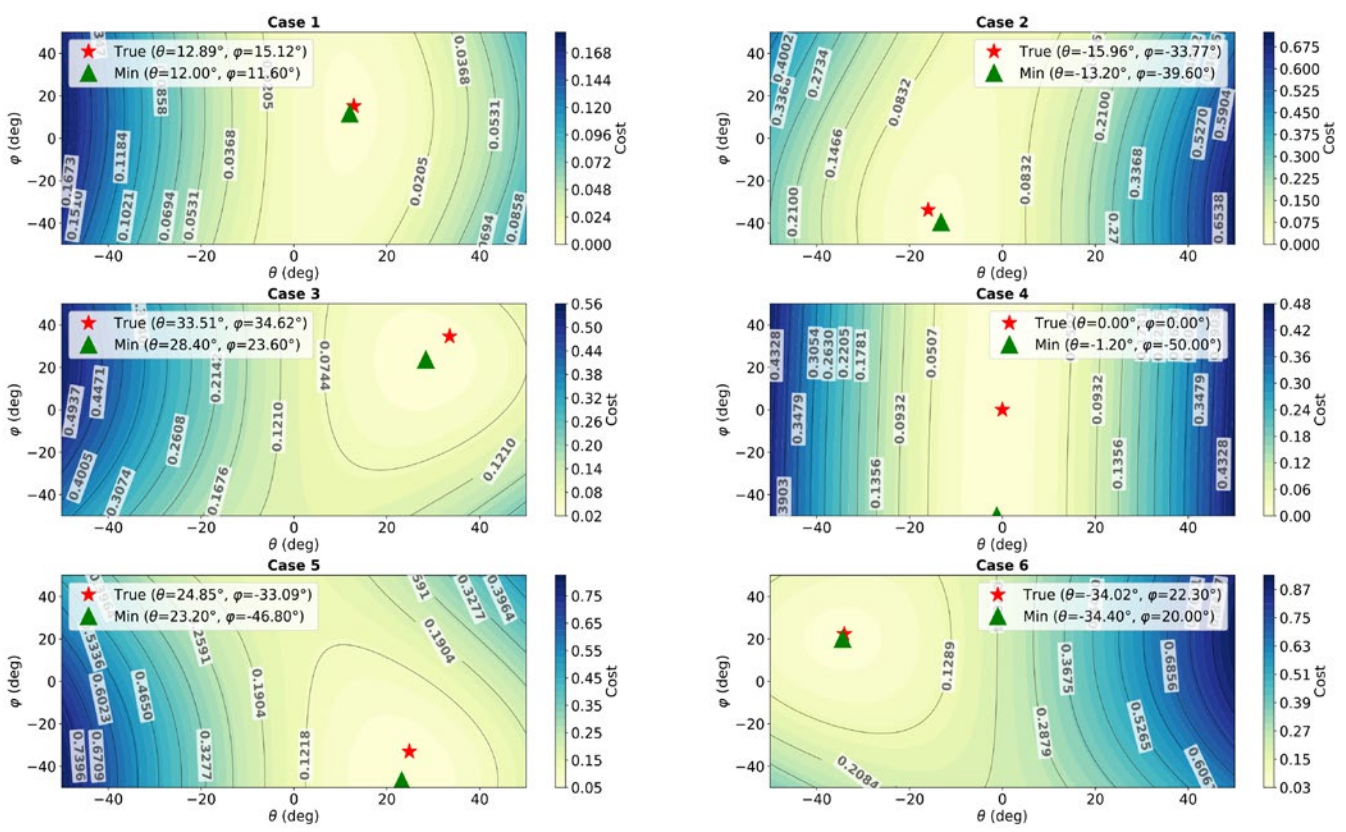

Fig. 3. Two-dimensional cost function landscapes for joint estimation of $\theta$ and $\varphi$ under different configurations. The red stars denote the ground-truth parameters, while the green triangles indicate the minimum-cost solutions.

### *B. D-H-Based Visual Reprojection Factor*

For the considered wheeled humanoid robot, the visual reprojection residual of a feature $\mathbf{p}_k$ between the *i*-th and *j*-th image frames can be formulated as

$$\mathbf{r}_{ij} = \mathbf{R}_{I_j}^{C_j} \left\{ \underbrace{\mathbf{R}_W^{I_j} \left[ \mathbf{R}_{I_i}^{W} \left( \mathbf{R}_{C_i}^{I_i} \underbrace{\frac{1}{\mu_k} \bar{\mathbf{p}}_k}_{\mathbf{p}_{k,i}} + \mathbf{t}_{C_i}^{I_i} \right) + \mathbf{t}_{I_i}^{W} - \mathbf{t}_{I_j}^{W} \right]}_{\mathbf{p}_{k,j}} - \mathbf{t}_{C_j}^{I_j} \right\} - \tilde{\mathbf{p}}_j . \quad (3)$$

Unlike conventional visual-inertial systems that assume a constant camera-IMU extrinsic transformation, the proposed system considers the time-varying extrinsic parameters induced by the robot joint motions. As discussed in the previous section, the camera-IMU transformations at the *i*-th and *j*-th frames are different and denoted as $\mathbf{T}_{C_i}^{I_i}$ and $\mathbf{T}_{C_j}^{I_j}$, respectively. Therefore, the reprojection residual incorporates the corresponding dynamic extrinsic transformations at both image frames. Based on (3), the Jacobian of the visual residual with respect to the system state variables $\boldsymbol{x}$:

$$\frac{\partial \mathbf{r}_{ij}}{\partial \boldsymbol{x}} = \left[ \frac{\partial \mathbf{r}_{ij}}{\partial \mathbf{T}_{I_i}^{W}} \quad \frac{\partial \mathbf{r}_{ij}}{\partial \mathbf{T}_{I_j}^{W}} \quad \frac{\partial \mathbf{r}_{ij}}{\partial \varphi_i} \quad \frac{\partial \mathbf{r}_{ij}}{\partial \varphi_j} \quad \frac{\partial \mathbf{r}_{ij}}{\partial \theta_i} \quad \frac{d\mathbf{r}_{ij}}{d\theta_j} \quad \frac{d\mathbf{r}_{ij}}{d\mu_i} \right], \quad (4)$$

where

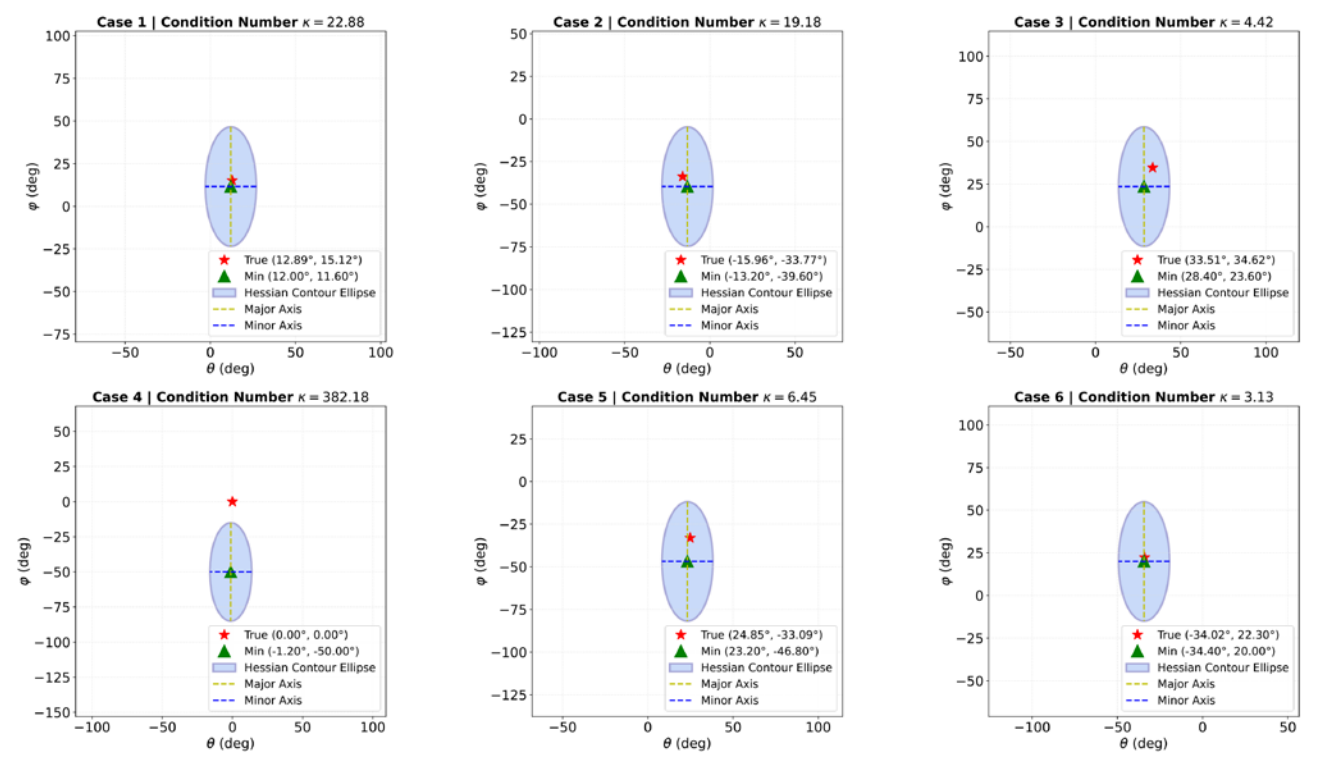

Fig. 4. Local Hessian analysis of $\theta$ and $\varphi$. The ellipse represents the local curvature of the cost function around the minimum, where the major and minor axes correspond to the eigenvectors associated with the minimum and maximum eigenvalues, respectively.

$$\begin{aligned} \frac{\partial \mathbf{r}_{ij}}{\partial \Delta_i} &= \mathbf{R}_{I_j}^{C_j} \mathbf{R}_W^{I_j} \mathbf{R}_{I_i}^{W} \left( \frac{\partial \mathbf{R}_{C_i}^{I_i}}{\partial \Delta_i} \mathbf{p}_{k,i} + \frac{\partial \mathbf{t}_{C_i}^{I_i}}{\partial \Delta_i} \right) \\ \frac{\partial \mathbf{r}_{ij}}{\partial \Delta_j} &= \frac{\partial \mathbf{R}_{I_j}^{C_j}}{\partial \Delta_j} \left( \mathbf{p}_{k,j} - \mathbf{t}_{C_j}^{I_j} \right) - \mathbf{R}_{I_j}^{C_j} \frac{\partial \mathbf{t}_{C_j}^{I_j}}{\partial \Delta_j}, \end{aligned} \quad (5)$$

where $\Delta$ denotes either $\varphi$ or $\theta$. In (4), the Jacobian of the residual with respect to the pose and inverse depth follow the formulation in [10]. The Jacobian of the camera-IMU extrinsic parameters with respect to the joint states can be directly derived from (2) as

$$\frac{\partial \mathbf{T}_C^I}{\partial \varphi} = \begin{bmatrix} c_\varphi c_\theta & 0 & -s_\varphi c_\theta & -L_3 s_\varphi c_\theta \\ s_\varphi & 0 & c_\varphi & L_3 c_\varphi \\ -c_\varphi s_\theta & 0 & s_\theta s_\varphi & L_3 s_\theta s_\varphi \\ 0 & 0 & 0 & 0 \end{bmatrix} = \begin{bmatrix} \frac{\partial \mathbf{R}_C^I}{\partial \varphi} & \frac{\partial \mathbf{t}_C^I}{\partial \varphi} \\ \mathbf{0}^T & 0 \end{bmatrix}, \quad (6)$$

and

$$\begin{aligned} \frac{\partial \mathbf{T}_C^I}{\partial \theta} &= \begin{bmatrix} -s_\varphi s_\theta & -c_\theta & -c_\varphi s_\theta & L_2 c_\theta - L_3 c_\varphi s_\theta \\ 0 & 0 & 0 & 0 \\ -s_\varphi c_\theta & s_\theta & -c_\theta c_\varphi & -L_2 s_\theta - L_3 c_\theta c_\varphi \\ 0 & 0 & 0 & 0 \end{bmatrix} \\ &= \begin{bmatrix} \frac{\partial \mathbf{R}_C^I}{\partial \theta} & \frac{\partial \mathbf{t}_C^I}{\partial \theta} \\ \mathbf{0}^T & 0 \end{bmatrix}. \end{aligned} \quad (7)$$

In (6) and (7), the Jacobians corresponding to the *i*-th and *j*-th keyframes share the same analytical expression.

Based on the visual reprojection residual in (3) and its corresponding Jacobian in (4), the proposed D-H-based visual reprojection factor can be formulated. Unlike conventional visual reprojection factors with fixed camera-IMU extrinsics, the proposed factor explicitly accounts for the time-varying extrinsic transformation during optimization, thereby maintaining consistency between the visual measurements and the robot kinematic model.

### *C. Coarse Angle Estimation*

Because the robot may start from an arbitrary configuration,

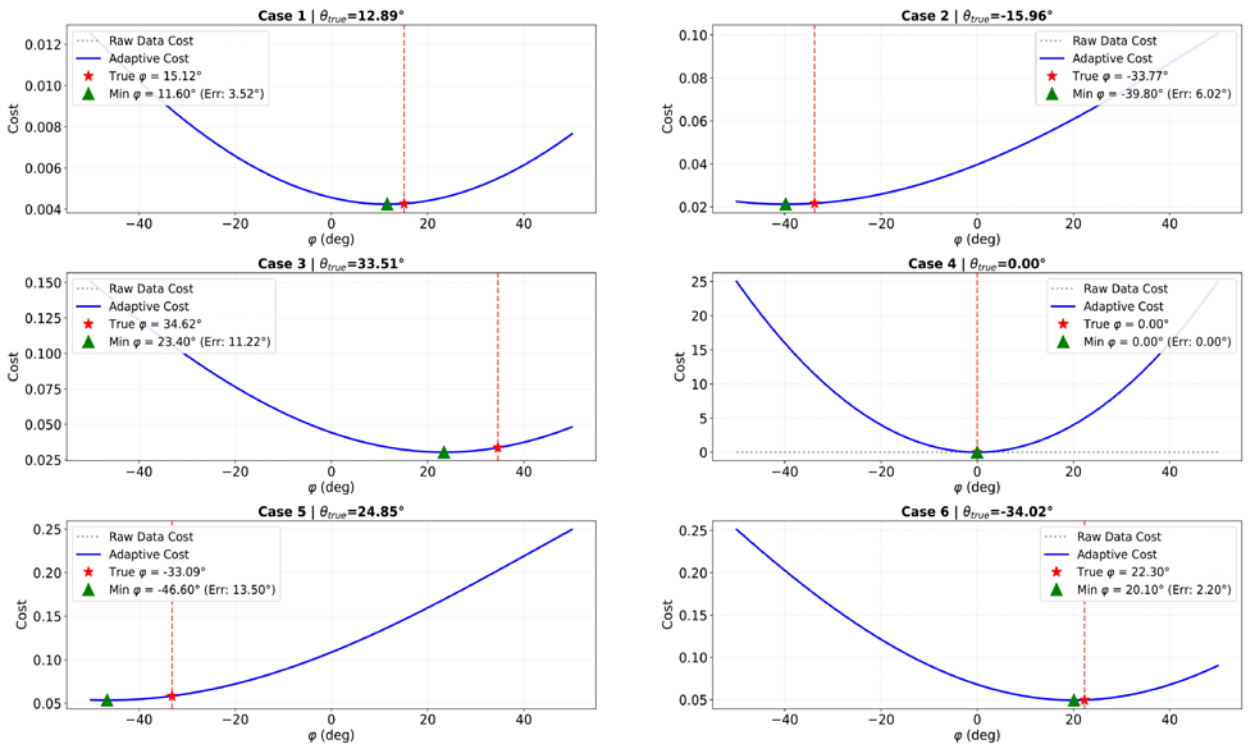

Fig. 5. Effect of adaptive regularization on $\varphi$ estimation. The original and regularized costs are shown by the gray dotted and blue solid curves, respectively, with the ground-truth and estimated values marked by red stars and green triangles.

both $\theta$ and $\varphi$ can deviate from zero at system startup. Reliable initial estimates of these two joint angles are therefore required to provide a suitable initialization for subsequent optimization and facilitate convergence. However, since the robot operates predominantly on a planar surface, the available rotational motion provides insufficient geometric constraints for jointly estimating the two parameters, making the optimization prone to degeneracy. To address this issue, we first analyze the observability and coupling characteristics of $\theta$ and $\varphi$, and then propose a two-step initialization strategy with adaptive regularization.

For a set of relative camera and IMU rotations $\left(\mathbf{R}_{C_i}^{C_{i+1}}, \mathbf{R}_{I_i}^{I_{i+1}}\right)$ and $i \in \{1, \cdots, N\}$, the logarithmic rotation residual cost function is defined as

$$J(\theta, \varphi) = \sum_{i=1}^{N} \left\| \mathrm{Log}^{\vee}\left(\mathbf{R}_{C_i}^{C_{i+1}T} \mathbf{R}_I^C \mathbf{R}_{I_i}^{I_{i+1}} \mathbf{R}_C^I\right) \right\|_2^2, \tag{8}$$

where $\mathrm{Log}^{\vee}(\mathbf{R}) = \alpha \mathbf{u}$ denotes the logarithm map.

We collected six datasets with different initial configurations and performed a grid search over $\theta$ and $\varphi$ to evaluate the cost landscape. The contour maps are shown in Fig. 3. The contour lines are significantly denser along the $\theta$ direction, indicating that the cost function changes more rapidly with respect to $\theta$ and is therefore more sensitive to this parameter. In contrast, the cost varies much more slowly along the $\varphi$ direction, indicating that the available measurements provide weaker constraints on $\varphi$. Therefore, although the cost function contains sufficient information to obtain a relatively accurate estimate of $\theta$, directly estimating $\varphi$ from the measurement residual alone is less reliable.

To further quantify the coupling and relative observability of the two parameters, we analyze the Hessian matrix of the cost function around its minimum. Let

$$\mathbf{H} = \nabla^2 J(\theta, \varphi). \tag{9}$$

The eigenvalues and eigenvectors of $\mathbf{H}$ characterize the local curvature of the cost function along different parameter directions. The corresponding equivalent uncertainty ellipses are shown in Fig. 4. The short axis of the ellipse is primarily aligned with the $\theta$ direction and corresponds to the larger eigenvalue, indicating a stronger constraint on $\theta$. Conversely, the long axis is mainly aligned with the $\varphi$ direction and corresponds to the smaller eigenvalue, indicating a weaker constraint on $\varphi$. Moreover, the principal axes of the ellipses are approximately aligned with the $(\theta, \varphi)$ coordinate axes, suggesting that the cross-coupling term of $\mathbf{H}$ is relatively small. Thus, $\theta$ and $\varphi$ exhibit an approximately decoupled local structure, motivating the decomposition of the joint optimization into two sequential estimation steps: first estimating $\theta$, followed by estimating $\varphi$.

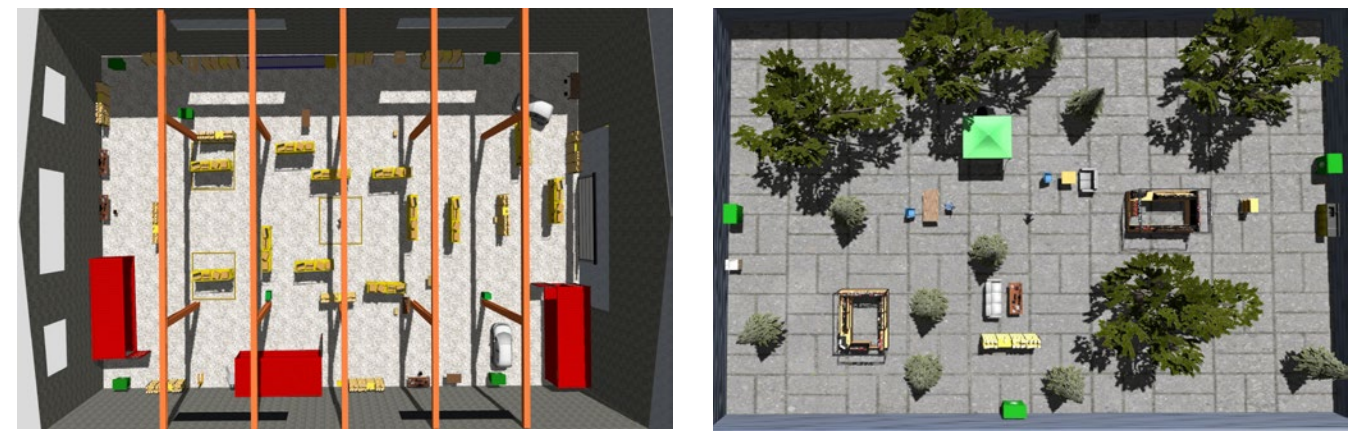

(a) (b)

Fig. 6. Gazebo simulation environments: (a) *Scene 1* (indoor industrial factory); (b) *Scene 2* (outdoor environment).

However, a more severe degeneracy occurs when $\theta$ approaches zero. In this configuration, the rotational motion induced by $\varphi$ becomes difficult to distinguish from the dominant planar rotational motion of the robot. Consequently, the cost function becomes nearly flat along the $\varphi$ direction, i.e.,

$$\frac{\partial J(\theta \to 0, \varphi)}{\partial \varphi} \approx 0. \tag{10}$$

In this degenerate case, the measurement-based cost provides insufficient information for determining a unique $\varphi$, and the estimated value may become highly sensitive to measurement noise or drift toward the boundary of the search interval. To improve the robustness of the initialization in this special case, we introduce an adaptive Gaussian regularization term into the $\varphi$ estimation process, as illustrated in Fig. 5.

Specifically, the proposed two-step strategy first fixes $\varphi = 0$ and estimates $\theta$ by minimizing $J(\theta, 0)$ over a predefined interval $[\theta_{\min}, \theta_{\max}]$. A one-dimensional golden-section search is adopted to obtain the coarse estimate $\hat{\theta}$. Subsequently, $\theta$ is fixed to $\hat{\theta}$, and $\varphi$ is estimated by minimizing the following regularized cost function:

$$\min_{\varphi} J_{total}(\varphi) = J(\hat{\theta}, \varphi) + \lambda(\hat{\theta})(\varphi - \varphi_0)^2, \tag{11}$$

where $\varphi_0$ denotes the nominal installation angle. In this work, the nominal value is set to $\varphi_0 = 0^{\circ}$. The adaptive regularization coefficient is defined as

TABLE I

QUANTITATIVE EVALUATION OF 10 COARSE ANGLE ESTIMATION TRIALS IN SCENE 2. ALL ANGLES ARE GIVEN IN DEGREES

| Sequences | $\theta$ | | | $\varphi$ | | |
|---|---|---|---|---|---|---|
| | GT | Estimated | Abs. Error | GT | Estimated | Abs. Error |
| Ang_01 | 12.777 | 12.147 | 0.630 | 11.860 | 9.568 | 2.292 |
| Ang_02 | 22.632 | 18.335 | 4.297 | 24.179 | 22.058 | 2.121 |
| Ang_03 | 30.367 | 23.147 | 7.220 | 15.756 | 14.782 | 0.974 |
| Ang_04 | -11.860 | -11.172 | 0.688 | 15.642 | 12.147 | 3.495 |
| Ang_05 | -20.913 | -19.595 | 1.318 | 5.157 | 4.183 | 0.974 |
| Ang_06 | -32.258 | -31.226 | 1.032 | 14.840 | 17.189 | 2.349 |
| Ang_07 | -27.101 | -22.918 | 4.183 | 35.065 | 32.945 | 2.120 |
| Ang_08 | -21.887 | -20.569 | 1.318 | -23.377 | -22.575 | 0.802 |
| Ang_09 | 19.882 | 14.381 | 5.501 | -33.919 | -35.294 | 1.375 |
| Ang_10 | 14.209 | 12.605 | 1.604 | -15.527 | -16.444 | 0.917 |

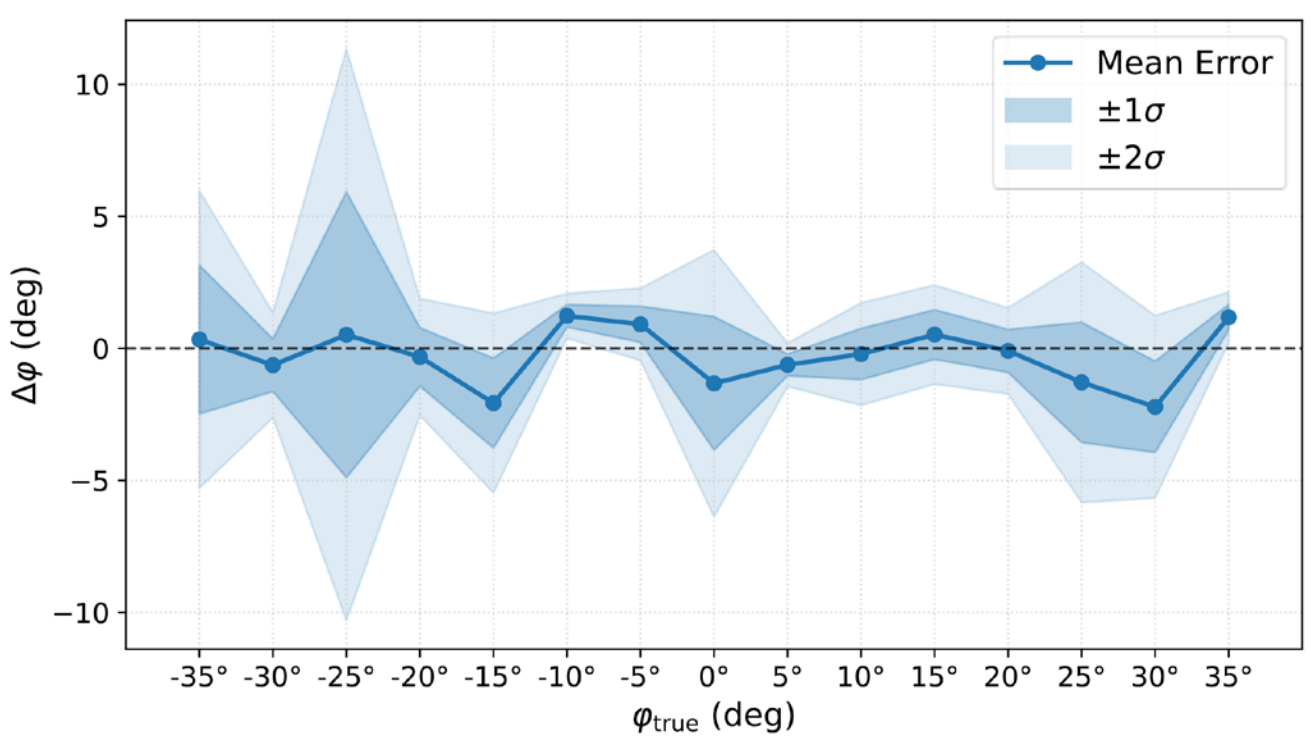

Fig. 7. Mean $\varphi$ estimation error with $1\sigma$ and $2\sigma$ uncertainty bands under the degenerate condition $\theta = 0^\circ$.

$$\lambda\left(\hat{\theta}\right) = \lambda_0 \exp\left(-\frac{\hat{\theta}^2}{2\sigma_\theta^2}\right), \tag{12}$$

where $\lambda_0$ is the maximum regularization strength and $\sigma_\theta$ controls the decay scale with respect to $\hat{\theta}$. When $\hat{\theta}$ approaches zero, the Gaussian kernel approaches its maximum value $\lambda\left(\hat{\theta}\right) \rightarrow \lambda_0$, and the quadratic prior term $\lambda\left(\hat{\theta}\right)\left(\varphi - \varphi_0\right)^2$ introduces a well-defined minimum around the nominal installation angle. This prevents the estimated $\varphi$ from drifting arbitrarily in the nearly flat cost landscape. Conversely, for a sufficiently large $|\hat{\theta}|$, e.g., $|\hat{\theta}| > 3\sigma_\theta$, the regularization coefficient rapidly approaches zero, and the total cost in (11) is dominated by the measurement-driven term. In this case, the algorithm relies primarily on the available rotational constraints to estimate $\hat{\varphi}$, thereby reducing the influence of the prior.

The proposed procedure therefore provides a coarse initialization of the two rotational extrinsic parameters by exploiting their approximately decoupled observability characteristics while explicitly handling the degenerate case around $\theta \rightarrow 0$. After initializing $\theta$ and $\varphi$, according to the D-H-based extrinsic model defined in (2), both the rotational and translational extrinsic parameters between the camera and IMU are uniquely determined by $\theta$ and $\varphi$. Therefore, the initial estimates of $\theta$ and $\varphi$ provide not only the initial rotational extrinsic parameters but also the corresponding translational extrinsic parameters, thereby completing the initialization of the full camera-IMU extrinsics.

## V. EXPERIMENTS

In this section, we evaluate the performance of the proposed modules and compare them with representative existing methods in Gazebo simulation environments and on multiple public datasets. All experiments are conducted on a computer equipped with an Intel Core Ultra 9 285H CPU at 2.90 GHz, 64 GB of RAM, and an NVIDIA GeForce RTX 5050 GPU.

### *A. Gazebo Configuration*

The simulation experiments are conducted in Gazebo using two representative environments, as shown in Fig. 6. The first is an indoor industrial factory environment with an approximate size of 50 m × 30 m, containing multiple shelves and other structural elements. The second is an outdoor environment with an approximate size of 40 m × 32 m, featuring various objects such as trees, stalls, and tents. The differential-drive wheeled humanoid robot described in Fig. 1 is deployed in both environments to evaluate the performance of the proposed framework and its individual components under different scene configurations.

The robot has a wheel separation of 0.50 m and a wheel radius of 0.08 m. A monocular camera and an IMU are mounted according to the configuration shown in Fig. 1. The IMU operates at 200 Hz, with zero-mean Gaussian white measurement noise having standard deviations of 0.01 rad/s and 0.01 $m/s^2$ for angular velocity and linear acceleration, respectively. In addition, temporally correlated dynamic biases are incorporated into both measurements to model slowly varying sensor drift.

The camera operates at 20 Hz with a resolution of 640 × 480 pixels and a horizontal field of view of 1.658 rad. Gaussian noise with a standard deviation of 0.007 is applied to the normalized image intensity. Wheel odometry is generated by the Gazebo differential-drive system based on the simulated wheel motion, while the robot pose provided by Gazebo is used as the ground truth for quantitative evaluation.

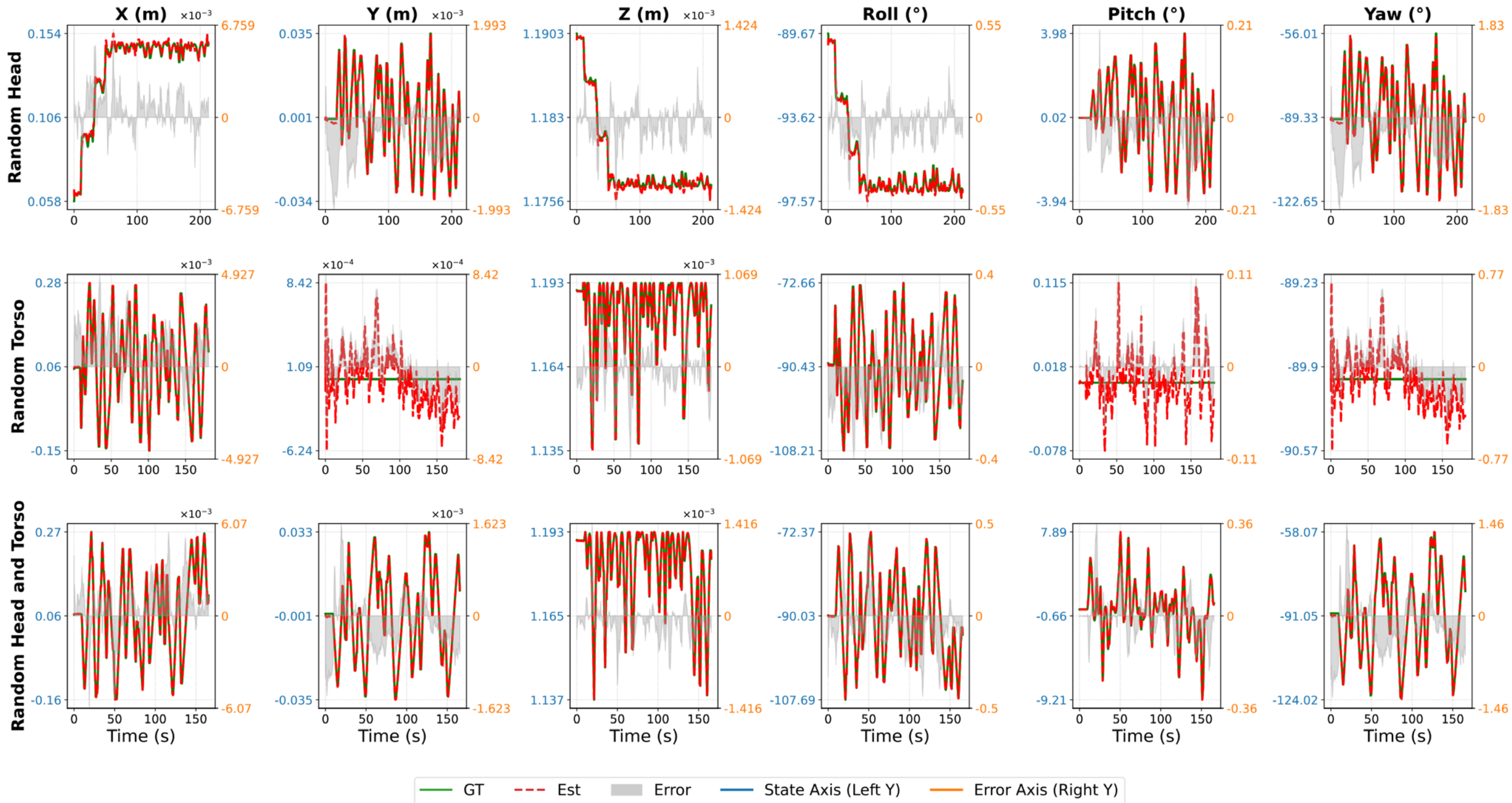

Fig. 8. Online dynamic extrinsic calibration results under three experimental configurations. In each subplot, the left y-axis represents the state variables (x, y, z, roll, pitch, yaw), while the right y-axis shows the corresponding estimation errors.

TABLE II
QUANTITATIVE COMPARISON OF CAMERA-IMU EXTRINSIC PARAMETER ESTIMATION ERRORS WITH FIXED EXTRINSICS

| Method | Translation (m) | | Rotation (deg) | |
|---|---|---|---|---|
| | RMSE | Mean | RMSE | Mean |
| VINS-Mono | 0.5625 | 0.3972 | **0.9040** | 0.8175 |
| Ours | **0.0022** | **0.0018** | 0.9047 | **0.7547** |

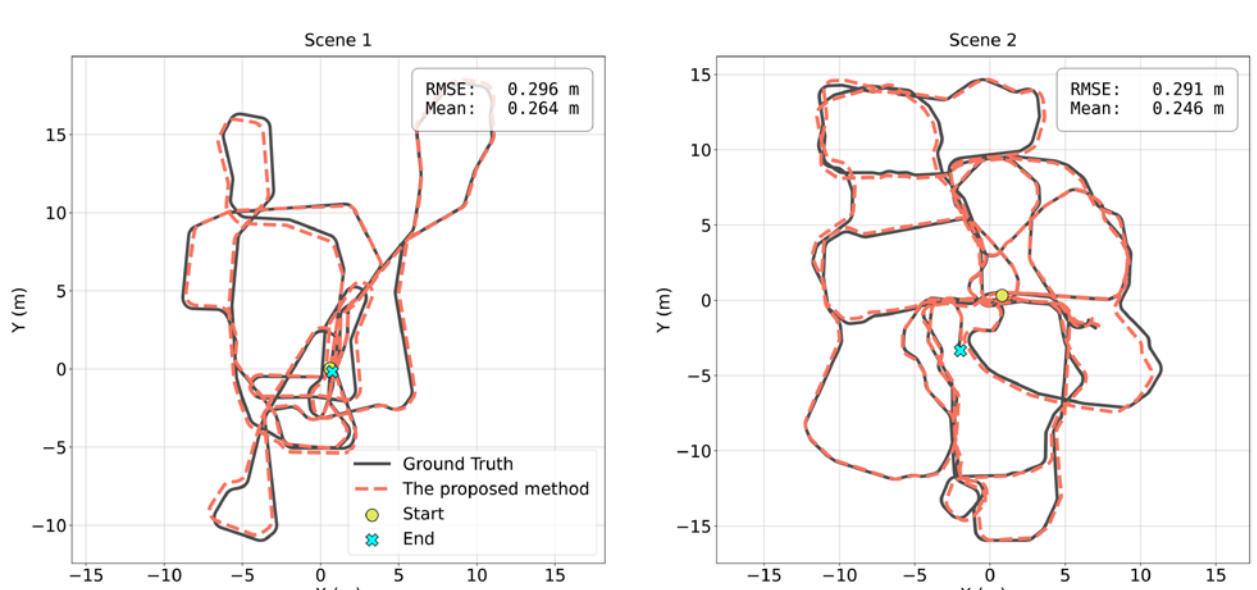

Fig. 9. Localization results under dynamic camera-IMU extrinsic parameters in two scenes.

During the experiments, the torso and head joint angles are varied within $\theta \in [-0.3,\ 0.3]$ rad and $\varphi \in [-0.6,\ 0.6]$ rad, respectively, unless otherwise specified. A relatively limited torso-motion range is adopted in the general experiments because excessive pitch motion may orient the camera toward the ground or sky, thereby substantially reducing the number of observable visual features. To evaluate the robustness of the proposed joint-angle initialization method under more challenging configurations, a larger torso-motion range is additionally considered in the corresponding experiments.

### *B. The Evaluation of Coarse Angle Estimation*

In this experiment, both $\theta$ and $\varphi$ are searched within the range of $[-50^\circ, 50^\circ]$. Ten different configurations of $\theta$ and $\varphi$ are evaluated in *Scene 2* to assess the proposed two-step angle initialization strategy, with the results summarized in Table I. For each configuration, the ground-truth (GT) angle, estimated angle, and corresponding absolute error are reported for both $\theta$ and $\varphi$.

For $\theta$, the absolute estimation error ranges from 0.630° to 7.220°, with an average error of approximately 2.8°. Such an error level provides a suitable initial estimate for the subsequent nonlinear optimization. The largest error is observed in Ang_03, where the ground-truth torso angle is approximately 30°. At this configuration, the camera is predominantly oriented toward the ground, resulting in visual features that are concentrated on relatively planar regions. The resulting visual observations provide less informative geometric constraints for estimating the inter-frame camera rotation, which in turn affects the accuracy of the estimated torso angle. In contrast, Ang_06 has a ground-truth torso angle of approximately -32°, while its absolute estimation error is only 1.032°. In this configuration, the camera is oriented more toward the upper part of the scene, where features can still be observed on structures such as trees and the upper portions of stalls. These observations provide richer geometric information for estimating the camera motion, leading to a more accurate torso-angle initialization. This comparison indicates that the accuracy of $\theta$ estimation is influenced not only by the magnitude of the torso rotation but also by the visual structure observed under the corresponding camera orientation.

The estimation of $\varphi$ exhibits smaller errors than that of $\theta$. The maximum absolute error is below 3.5°, while the average error is approximately 1.74°. The relatively accurate estimation of $\varphi$ can be attributed to the proposed two-step estimation strategy. Specifically, the first step provides a coarse estimate of $\theta$, which is subsequently used to constrain the estimation of $\varphi$ in the second step. As a result, the search for $\varphi$ can be performed based on a more informative estimate of the torso orientation. Moreover, the estimation accuracy of $\varphi$ remains relatively stable even when its magnitude exceeds 30°. This is mainly because head rotation changes the viewing direction while maintaining a substantial portion of the surrounding scene within the camera's field of view, allowing a sufficient number of visual features to remain observable.

Overall, the results in Table I demonstrate that the proposed two-step strategy provides reasonably accurate initial estimates for both $\theta$ and $\varphi$ across different joint configurations. The obtained initialization errors are sufficiently small to provide a practical starting point for the subsequent nonlinear optimization.

### *C. Degeneracy Analysis of Angle Initialization*

To investigate the behavior of the proposed angle initialization strategy under the degenerate condition $\theta = 0^\circ$, 15 configurations are evaluated by varying $\varphi$ from −35° to 35° at 5° intervals. For each configuration, the estimation error of $\varphi$ is recorded from the initialization stage through the subsequent backend optimization. The mean estimation error is then computed for each $\varphi$ configuration, together with the corresponding $\pm 1\sigma$ and $\pm 2\sigma$ error bands, as shown in Fig. 7. The mean estimation error of $\varphi$ remains within approximately $\pm 3^\circ$ across the tested range, with the largest mean errors occurring at $\varphi = 30^\circ$ and 35°. The $\pm 1\sigma$ error bands generally remain within approximately $\pm 5^\circ$, while the $\pm 2\sigma$ error bands are mostly within approximately $\pm 6^\circ$, except for the case at $\varphi = -25^\circ$, where a relatively larger deviation is observed. This non-monotonic behavior is likely related to changes in the observable scene structure and visual feature distribution under different head orientations.

The observed error bands are primarily attributed to the uncertainty in the initial $\varphi$ estimates. During the subsequent backend optimization, the estimated $\varphi$ remains relatively stable without noticeable deviation from its final optimized value. This indicates that, although the degenerate configuration may introduce some uncertainty into the initial estimation of $\varphi$, the subsequent optimization can still maintain a relatively stable estimate. Overall, these results suggest that setting $\varphi = 0^\circ$ for initialization under the degenerate condition $\theta = 0^\circ$ does not introduce noticeable degradation in the backend optimization over the tested configurations. This supports the use of the proposed adaptive regularization strategy for handling the angle-initialization degeneracy.

### *D. Online Dynamic Extrinsic Parameter Calibration*

The performance of the proposed D-H-based online dynamic extrinsic calibration method is evaluated in three motion configurations, as shown in Fig. 8. The first row corresponds to random head motion, i.e., variations in $\varphi$. Since $\theta = 0^\circ$ leads to a degenerate configuration, a small initial torso rotation is introduced to avoid this condition. The second row considers random torso motion, i.e., variations in $\theta$, while the third row considers simultaneous random motions of both the head and torso. For each configuration, the six columns show the variations of the three translational and three rotational extrinsic parameters, respectively.

As shown in Fig. 8, the estimation errors of the three translational extrinsic parameters remain at the millimeter level across all three motion configurations, indicating that the proposed D-H-based method can track the dynamic translational extrinsic parameters online with relatively small errors. For the rotational extrinsic parameters, when only the torso undergoes random motion, the ground-truth pitch and yaw remain constant. Nevertheless, small variations can be observed in their estimated values because the backend continues to jointly optimize the complete extrinsic parameters. The corresponding angular errors remain within 1°, suggesting that these variations have a limited effect on the overall localization performance. The simultaneous motion of the head and torso represents the most challenging configuration. Nevertheless, both the translational and rotational estimates remain close to their corresponding ground-truth values. The roll and pitch errors remain within 0.5°, while the yaw error is relatively larger. The maximum yaw error of approximately 1.5° occurs shortly after the initialization stage, after which the error remains generally within 1° as the backend optimization becomes stable. Overall, these results demonstrate that the proposed method can continuously estimate the dynamic camera-IMU extrinsic parameters under both individual and simultaneous head-torso motions, showing its robustness to dynamic sensor configurations.

Since our method is developed based on the VINS-Mono, we further compare the extrinsic parameter estimation performance of VINS-Mono and the proposed D-H-based method in terms of the mean error and RMSE of the translational and rotational components, as reported in Table II. Since VINS-Mono does not explicitly account for dynamically varying extrinsic parameters, the camera-IMU extrinsic parameters are kept fixed throughout this experiment to provide a fair comparison. As shown in Table II, the proposed method achieves substantially lower errors in the translational extrinsic parameters than VINS-Mono. This improvement can be attributed to the D-H-based modeling of the robot's actual multi-link mechanical structure, which introduces explicit structural constraints on the camera-IMU transformation. In particular, the proposed extrinsic model in (2) constrains the translational parameters according to the underlying mechanical configuration, thereby reducing their deviation during optimization. For the rotational extrinsic parameters, the RMSE of the proposed method is comparable to that of VINS-Mono, while its mean error is lower. The relatively larger rotational error compared with the translational error can be partly attributed to the fixed configuration with $\theta = 0^\circ$ and $\varphi = 0^\circ$, which corresponds to the degenerate condition discussed previously. Nevertheless, the rotational errors of both methods remain below 1° on average, indicating that the resulting extrinsic estimates provide sufficiently accurate calibration for the localization task.

### *E. Localization under Time-Varying Camera-IMU Extrinsics*

The localization performance of the proposed method under dynamically varying camera-IMU extrinsic parameters is evaluated in two scenes, as shown in Fig. 9. During both experiments, the robot simultaneously performs random motions of the head and torso, resulting in continuously varying camera-IMU extrinsic parameters. The trajectory RMSE is 0.296 m for *Scene 1* and 0.291 m for *Scene 2*. As shown in Fig. 9, the estimated trajectories generally follow the corresponding ground-truth trajectories in both scenes. Combined with the dynamic extrinsic calibration results in *Section D*, these results indicate that the proposed online extrinsic estimation can provide sufficiently accurate time-varying extrinsic estimates to support stable localization under simultaneous head and torso motions.

## VI. CONCLUSION

In this work, we address the challenges of time-varying camera-IMU extrinsic calibration in VIO systems for ground robots. A D-H-based camera-IMU extrinsic model is incorporated into the VIO backend to jointly optimize the robot joint angles and continuously estimate the time-varying extrinsic parameters. A two-step initialization strategy is further proposed to provide reliable initial estimates of the extrinsic parameters under different initial robot orientations. Experiments in Gazebo demonstrate that the proposed framework can effectively handle time-varying camera-IMU extrinsics and maintain stable localization performance for ground robots.